%% file: root.tex
\documentclass[letterpaper, 10 pt, conference]{ieeeconf}  

\IEEEoverridecommandlockouts                              

\usepackage{epsfig} 
\usepackage{times} 
\usepackage{amsmath} 
\usepackage{amssymb}  

\include{include/commands}
\include{include/packages}

\title{\LARGE \bf
FloMo: Tractable Motion Prediction with Normalizing Flows
}

\author{Christoph Sch\"oller$^{1}$ and Alois Knoll$^{2}$
\thanks{$^{1}$Christoph Sch\"oller is with fortiss GmbH, Research Institute of the Free State of Bavaria, Munich, Germany}%
\thanks{$^{2}$Alois Knoll is with the Technical University of Munich, Munich, Germany}%
}

\begin{document}

\maketitle

\begin{abstract}
\input{sections/abstract}
\end{abstract}

\section{Introduction}
\label{sec:intro}
\input{sections/introduction}

\section{Related Work}
\label{sec:related}
\input{sections/related_work}

\section{Problem and Notation}
\label{sec:problem}
\input{sections/problem_formulation}

\section{Method}
\label{sec:method}
\input{sections/method}

\section{Experiments}
\label{sec:experiments}
\input{sections/experiments}

\section{Conclusion}
\label{sec:conclusion}
\input{sections/conclusion.tex}

\section*{Acknowledgment}
This research was funded by the Federal Ministry of Transport and Digital Infrastructure of Germany in the project \mbox{Providentia++}.

{\small
\bibliographystyle{IEEEtran}
\bibliography{include/references}
}

\end{document}

%% file: include/commands.tex
\newcommand{\fig}[1]{Fig.~\ref{#1}}
\newcommand{\sect}[1]{Sec.~\ref{#1}}
\newcommand{\tab}[1]{Tab.~\ref{#1}}
\newcommand{\eq}[1]{Eq.~\ref{#1}}

\newcommand{\obs}{\mathbf{o}}
\newcommand{\pred}{\mathbf{x}}
\newcommand{\predelem}{x}
\newcommand{\pos}{\mathbf{p}}
\newcommand{\obsenc}{\mathbf{c}}
\newcommand{\traj}{\bm{\phi}}
\newcommand{\bfx}{\mathbf{x}}
\newcommand{\bfu}{\mathbf{u}}
\newcommand{\g}{\,|\,}
\newcommand{\tighteq}{\mkern2.0mu{=}\mkern2.0mu}
\newcommand{\medeq}{\mkern3.0mu{=}\mkern3.0mu}

\newcommand{\setbr}[1]{\left\{#1\right\}}
\newcommand{\roundbr}[1]{\left(#1\right)}
\newcommand{\cornerbr}[1]{\left[#1\right]}

%% file: include/packages.tex
\usepackage{graphicx}
\usepackage{bm}
\usepackage{footnote}
\usepackage{hhline}
\usepackage{subcaption}
\usepackage{color}
\usepackage{siunitx}
\usepackage{nicefrac}

%% file: sections/abstract.tex
The future motion of traffic participants is inherently uncertain. To plan safely, therefore, an autonomous agent must take into account multiple possible trajectory outcomes and prioritize them. Recently, this problem has been addressed with generative neural networks. However, most generative models either do not learn the true underlying trajectory distribution reliably, or do not allow predictions to be associated with likelihoods. In our work, we model motion prediction directly as a density estimation problem with a normalizing flow between a noise distribution and the future motion distribution. Our model, named FloMo, allows likelihoods to be computed in a single network pass and can be trained directly with maximum likelihood estimation. Furthermore, we propose a method to stabilize training flows on trajectory datasets and a new data augmentation transformation that improves the performance and generalization of our model. Our method achieves state-of-the-art performance on three popular prediction datasets, with a significant gap to most competing models.

%% file: sections/introduction.tex
For autonomous agents like vehicles and robots, it is essential to accurately predict the movement of other agents in their vicinity. Only with this ability collisions can be avoided and interactions become safe. However, trajectories can never be predicted with absolute certainty and multiple future outcomes must be taken into account.

To address this problem, research on generative models for motion prediction has recently gained attention. An ideal generative model is expressive and able to learn the true underlying trajectory distribution. Furthermore, it allows the assignment of a likelihood value to each prediction. The knowledge of how likely certain trajectories are is important to prioritize, because it is infeasible for an agent to take into account all possible future behaviors of surrounding agents.

Yet, most methods do not have all of these desirable properties. For example, Generative Adversarial Networks (GANs) have been used extensively for motion prediction~\cite{gupta2018social, sadeghian2018sophie, socialways2019amirian}, but suffer from mode collapse and are not guaranteed to learn the true distribution of the data~\cite{salimans2016improved, che2016mode}. Variational Autoencoders (VAEs) are a popular type of generative models as well~\cite{lee2017desire, agarwal2020imitative, bhattacharyya2019conditional, xue2020scene} and approximate the true distribution with a lower bound. Unfortunately, likelihoods cannot be calculated directly with VAEs and must be estimated with computationally expensive Monte Carlo methods. Other contributions try to overcome the problem of missing likelihoods with the use of parametric density functions, most commonly normal distributions~\cite{casas2019spatially,  luo2020probabilistic}. However, this often requires unrealistic independence assumptions and provides only limited expressive power.
\begin{figure}[t]
    \centering
    \includegraphics[width=0.48\textwidth]{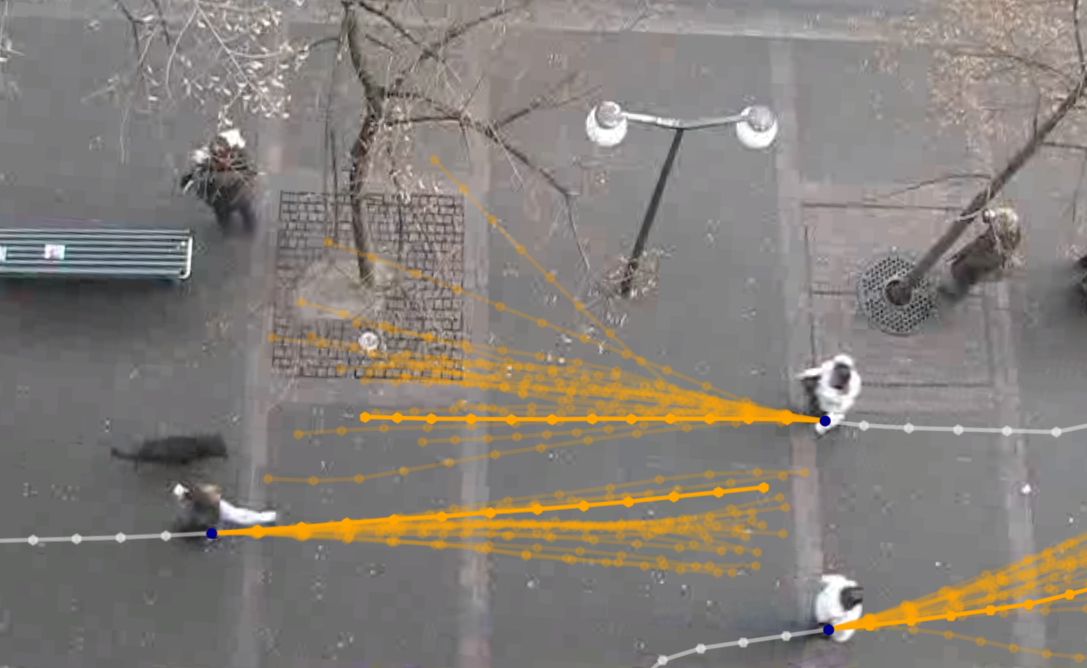}
    \caption{Trajectory predictions of our model (orange). More likely trajectories are drawn more opaque. The distributions our model learned are highly multi-modal.}
    \label{fig:cover-image}
\end{figure}

In this work, we propose a novel motion prediction model that addresses the aforementioned issues. In particular, our model FloMo is based on normalizing flows that we condition on observed motion histories. It is expressive and able to learn complex multi-modal distributions over future trajectories (see~\fig{fig:cover-image}). With FloMo, trajectories can be efficiently sampled and likelihoods are computed in closed form. These tractable likelihoods allow us to train our model with maximum likelihood estimation, instead of a proxy loss. Because, as we show, trajectory data is prone to cause divergence of likelihoods during training, we apply a novel noise injection method that significantly stabilizes training and enables the use of our model's likelihoods in downstream tasks. Furthermore, we propose a new data augmentation transformation that helps our model to generalize better and improves its performance. We demonstrate with an extensive evaluation on three popular motion prediction datasets that our method achieves state-of-the-art performance and we show, both qualitatively and quantitatively, that the likelihoods our model produces are meaningful.

%% file: sections/related_work.tex
Many classic approaches have been developed to make trajectory predictions~\cite{helbing1995social, pellegrini2009eth, pellegrini2011predicting}, and are still relevant today~\cite{scholler2020constant}.

\textbf{Neural Networks.} However, after successes on various other computer vision problems, neural networks have become popular for motion prediction as well. Alahi et al.~\cite{alahi2016social} use Long Short-Term Memories (LSTMs) to predict pedestrian trajectories and share information between agents with a social hidden state pooling. Similarly, Pfeiffer et al.~\cite{pfeiffer2018data} provide an LSTM with an occupancy grid of static objects and an angular grid of surrounding pedestrians. But also Convolutional Neural Networks (CNNs)~\cite{nikhil2018convolutional}, spatio-temporal graphs~\cite{vemula2018socialatt} or state refinement modules~\cite{zhang2019sr} have been proposed to predict single trajectories.

\textbf{Generative Models.} To predict not only a single trajectory, but multiple possible outcomes, prediction methods based on generative neural networks have been developed. Sadeghian et al.~\cite{sadeghian2018sophie} as well as Gupta et al.~\cite{gupta2018social} utilize GANs that are provided with additional context information. To fight mode collapse, Amirian et al.~\cite{socialways2019amirian} use an Info-GAN with an attention pooling module. The Trajectron++ model of Salzmann et al.~\cite{salzmann2020trajectron++} combines a conditional VAE, LSTMs and spatio-temporal graphs to produce multi-modal trajectory predictions. Inspired by BERT, Giuliari et al.~\cite{giuliari2020trajtransformer} propose to use a transformer architecture for motion prediction. Xue et al.~\cite{xue2020scene} propose the Scene Gated Social Graph that models the relations between pedestrians with a dynamic graph that is used to condition a VAE. Mohamed et al.~\cite{mohamed2020social} model social interactions with a spatio-temporal graph on which they apply graph convolutions and a temporal CNN to make predictions. Instead of directly predicting trajectories, Mangalam et al.~\cite{mangalam2020not} use a conditional VAE to first predict trajectory endpoints and a recursive social pooling module to make trajectory predictions. The prediction model of Pajouheshgar et al.~\cite{pajouheshgar2018back} is fully convolutional and outputs a discrete probability distribution over image pixels.

\textbf{Normalizing Flows.} While originally developed for density estimation~\cite{tabak2013family}, normalizing flows have recently been applied to various data generation problems~\cite{kim2018flowavenet, kingma2018glow}. In the area of motion prediction, normalizing flows have been rarely used. To generate trajectories for a planner, Agarwal et al.~\cite{agarwal2020imitative} sample from a conditional $\beta$-VAE~\cite{higgins2016beta} that uses a Neural Autoregressive Flow~\cite{huang2018neural} as a flexible posterior. Bhattacharyya et al.~\cite{bhattacharyya2019conditional} use a conditional Flow VAE with condition and posterior regularization to predict trajectories. In their recently published work~\cite{bhattacharyya2020haar}, they use a Block Autoregressive Flow based on Haar wavelets to learn distributions for motion prediction and also adapted FlowWaveNet~\cite{kim2018flowavenet} for motion prediction. Ma et al.~\cite{ma2020diverse} recently showed how to find those trajectories sampled from affine flows that are both likely and diverse to make predictions.

The method we propose in this work is a flow-based generative model that can learn complex multimodal distributions. It allows tractable likelihood computation and can be trained directly with maximum likelihood estimation. Most existing generative models only possess some of these properties. In contrast to the flow-based prediction models proposed in concurrent works~\cite{bhattacharyya2020haar, ma2020diverse}, the flow we use is based on splines and hence is more flexible, which our results demonstrate. Furthermore, we propose a novel noise injection method that significantly stabilizes training and a data augmentation transformation that further improves our model's generalization and performance. In our extensive experiments we show that our model achieves state-of-the-art results on popular motion prediction datasets and that the likelihoods it produces are meaningful and can be used to modulate how concentrated our model's predictions are.

%% file: sections/problem_formulation.tex
The motion of an agent can be defined as a finite sequence $\traj \medeq (\pos^0,...,\pos^T)$ of positions $\pos^t \medeq (x^t, y^t)$ over discrete timesteps $t\in\setbr{0,...,T}$. For predicting the future motion $\pred \medeq (\pos^{t+1},...,\pos^{t+n})$ of an agent, only a part $\obs \medeq (\pos^0,...,\pos^t)$ of its past trajectory is observable. From the perspective of generative modeling, the goal is to learn the conditional distribution $p(\pred\g\obs)$. Future trajectories can then be predicted by sampling $\hat{\pred} \sim p(\pred\g\obs)$.

\medskip

One way to learn such a distribution is to use normalizing flows. Normalizing flows are probabilistic models that can learn complex data distributions by transforming noise samples $\bfu$ from a simple base distribution $p_u(\bfu)$ into samples $\bfx$ from the target distribution:
\begin{equation}
\label{eq:flowtransform}
    \bfx = f(\bfu) \quad\text{where}\quad \bfu \sim p_u(\bfu).
\end{equation}
By defining the transformation $f(\bfu)$ such that it is invertibe and differentiable, the probability density of $\bfx$ can be obtained by a change of variables~\cite{papamakarios2019normalizing}:
\begin{equation}
\label{eq:flowforward}
    p_x(\bfx) = p_u(\bfu) {\left| \det \thinspace J_{f}(\bfu) \right|}^{-1}.
\end{equation}
Here $J_{f}(\bfu)$ denotes the Jacobian matrix of the function $f(\bfu)$. In the same manner, by the inverse function theorem, it is also possible to express $p_x(\bfx)$ in terms of $\bfx$ and $J_{f^{-1}}$:
\begin{equation}
\label{eq:flowinverse}
    p_x(\bfx) = p_u(f^{-1}(\bfx)) \left| \det \thinspace J_{f^{-1}}(\bfx) \right|.
\end{equation}
For the base distribution, usually a standard normal is chosen and the invertible transformation is implemented by a neural network. To make the flow more flexible, several such transformations can be composed. It is important that the Jacobian determinant can be computed efficiently and, depending on the use case, the flow must be easy to invert. Furthermore, to represent complex distributions the transformations in the flow must be expressive.

%% file: sections/method.tex
\begin{figure*}[t]
    \centering
    \includegraphics[width=1.0\textwidth]{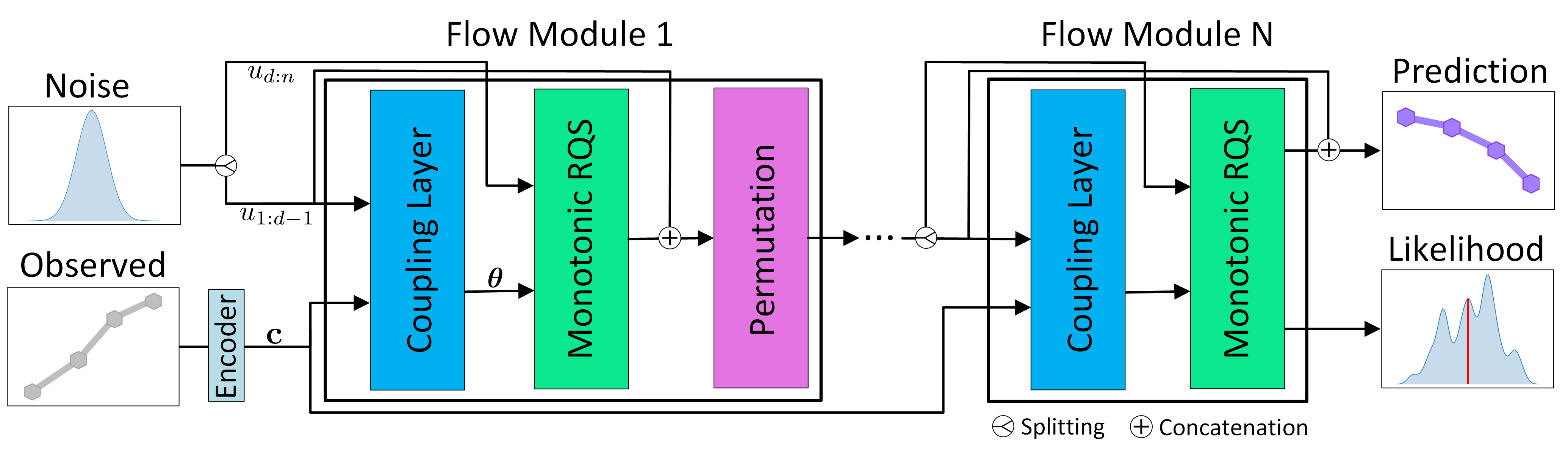}
    \caption{Our model is composed of multiple flow modules, each containing a coupling layer for conditioning, a monotonic rational-quadratic spline (RQS) transformation and -- except the last module -- a permutation layer. It receives an encoded observed trajectory and a noise vector, and outputs a prediction sample along with its likelihood.}
    \label{fig:architecture}
\end{figure*}

The objective of our model is to learn the conditional motion distribution $p(\pred\g\obs)$, where $\obs$ is an observed trajectory and $\pred$ is the trajectory to predict (see~\sect{sec:problem}). We learn this distribution by utilizing normalizing flows. To then make a prediction, we sample from a standard normal base distribution $\bfu\sim\mathcal{N}(\mu\tighteq0,\sigma\tighteq1)$ and pass the sample through our model, which we condition with the encoded observed trajectory $\obs$. The output of our model is a sampled trajectory prediction $\hat{\pred}$. By evaluating~\eq{eq:flowforward}, we can directly compute the likelihood of each sample in the same network pass. An overview of our architecture is given in~\fig{fig:architecture}. The main components of our model are a motion encoder and neural spline flows as proposed by Durkan et al.~\cite{durkan2019neural}, consisting of conditional coupling layers~\cite{dinh2014nice} and monotonic spline transformations~\cite{gregory1982piecewise}.

In this work we focus on the prediction of individual agents, because tests with integrating common interaction modules in our model's conditioning did not lead to relevant performance improvements. This is in line with the findings in~\cite{scholler2020constant,gupta2018social} and \cite{giuliari2020trajtransformer}. In the following sections, we explain each component of our model in detail, including how we prepare our data to achieve stable training, our objective function, and a novel trajectory augmentation transformation that we apply to increase generalization and performance.

\subsection{Motion Encoder}
\label{sec:motionencoder}
The first module of our model is the motion encoder, which encodes the observed trajectory $\obs$. Before we encode $\obs$, we subtract from each position $\pos^t \in \obs$ its preceding position, i.e. $ \pos^{\prime t} \medeq \pos^t - \pos^{t-1}$. This means instead of encoding absolute coordinates, we encode relative displacements, which has proven to be beneficial for motion prediction~\cite{becker2018red, scholler2020constant}. From now on, we will denote the resulting relative observed trajectory as $\obs^\prime$ and its encoding as $\obsenc$. We implement the encoder as a recurrent neural network with three Gated Recurrent Units (GRUs)~\cite{cho2014learning} and a hidden state size of 16. Before we pass each displacement step to the encoder, we embed it with a linear layer in a 16 dimensional vector. The output of the last GRU is then passed through an Exponential Linear Unit (ELU)~\cite{clevert2015fast} and again linearly transformed, while keeping 16 output dimensions. We determined these hidden and embedding sizes empirically. Because the ELU function is non-zero everywhere, it helps to avoid dying neurons in the network recursion. The recurrent architecture of our encoder enables it to work with input trajectories of various lengths.

\subsection{Conditional Coupling Layer}
\label{sec:coupling}
One way to design a normalizing flow is to modularize it into a transformation and a conditioner~\cite{huang2018neural}. The conditioner takes the input $\bfu$ and parameterizes the transformation that in turn transforms $\bfu$ into $\bfx$. In our work, it is important that our flow is fast to evaluate both in the forward and inverse direction. For sampling trajectories, we must transform forward from $\bfu$ to $\pred$, but during training we have to compute the likelihood of $\pred$ in the inverse direction (\eq{eq:flowinverse}). Furthermore, it would be desirable for our model to allow the computation of likelihoods for trajectories that an agent could possibly take, but that were not sampled. This also requires the inverse direction.

For the flow to be fast to invert, both the transformation and conditioner must be fast to invert. To achieve this for the conditioner, we use coupling layers~\cite{dinh2014nice, durkan2019neural} to implement our flow. Coupling layers are just as fast to invert, as they are to compute forward. Our coupling layer computes the output $\bfx$ as follows ($\oplus$ denotes concatenation):
\begin{equation}\begin{split}
\label{eq:coupling}
x_{1:d-1} = u_{1:d-1} \\
\bm{\theta} = \operatorname{NN}(u_{1:d-1} \oplus \obsenc) \\
x_i = \tau(u_i;\bm{\theta}_i)\text{ for i $\geq$ d}. \\
\end{split}\end{equation}
First, we split the input $\bfu$ in half and assign the first part $u_{1:d-1}$ directly to the output. Then we concatenate $u_{1:d-1}$ with trajectory encoding $\obsenc$ (see~\sect{sec:motionencoder}) and feed it to the conditioner network that computes the parameters $\bm{\theta}$. Using $\bm{\theta}$ to parameterize the invertible transformation $\tau$, we transform the second half $u_{d:n}$ of $\bfu$ element-wise to the remaining corresponding outputs. The resulting Jacobian matrix is lower triangular, and hence its determinant can be easily computed as the product of its diagonal elements~\cite{dinh2014nice}. By concatenating $\obsenc$ to the conditioner input, we make our flow conditional on the observed trajectory, such that it learns the density~$p(\pred\g\obs)$.

We implement the conditioner as a regular feed forward neural network with five hidden layers. Each layer has 32 neurons and is followed by an ELU activation. This configuration worked well empirically. Because half of the inputs are not transformed in a coupling layer, it is crucial to stack several such modules and randomly permute the input vectors between the modules. As permutations are volume-preserving, the Jacobian determinant of such a permutation layer is simply 1.

\subsection{Monotonic Spline Transforms}
\label{sec:neuralspline}
Transformations used in normalizing flows must be expressive, invertible and differentiable. In motion prediction, expressive power is crucial to represent complex distributions and only fast invertibility allows the computation of likelihoods for query trajectories at runtime and short training times. However, most expressive flows, e.g. neural flows~\cite{huang2018neural}, cannot be inverted analytically and we have to resort to iterative methods like bisection search~\cite{papamakarios2019normalizing}. On the other hand, flows that are fast to invert often use simple transformations, e.g. affine or linear transformations, and hence are not very expressive.

However, recently Durkan et al.~\cite{durkan2019neural} proposed using monotonic rational-quadratic splines (RQS)~\cite{gregory1982piecewise} as flow transformations. In conjunction with coupling layers, this kind of flow becomes both expressive and fast to invert. The spline transformation described in the following corresponds to the function $\tau$ in~\sect{sec:coupling}.

The spline is defined by $K$ different rational-quadratic functions that pass through $K\!+\!1$ knot coordinates $\setbr{\roundbr{x^k, y^k}}_{k=0}^K$. These knots monotonically increase between $\roundbr{x^0, y^0} \medeq (-B, -B)$ and $\roundbr{x^K, y^K} \medeq (B, B)$. In accordance with Durkan et al., we assign the spline $K - 1$ arbitrary positive derivatives $\setbr{\delta^k}_{k=1}^{K-1}$ for the intermediate knot connection points and set the boundary derivatives $\delta^0 \medeq \delta^K \medeq 1$ to match the linear \lq tails\rq~outside of the rational-quadratic support $[-B, B]$. This support is a hyper-parameter and is set manually. With these parameters, the spline is smooth and fully defined. The neural network that is parameterizing it can learn the knot positions and boundary derivatives during training.

The spline transformation is then applied element-wise, e.g. to a given scalar input $x_{in}$. If $x_{in}$ is outside the support, the identity transformation is applied. Otherwise, the correct knot bin is determined first, and then
\begin{equation}\begin{split}
s_k &= \roundbr{y^{k+1} - y^k}/\roundbr{x^{k+1} - x^k}\\
\xi &= \roundbr{x_{in} - x^k}/\roundbr{x^{k+1} - x^k}
\end{split}\end{equation}
are computed. After this, the forward transformation
\begin{equation}
\frac{\alpha^k(\xi)}{\beta^k(\xi)} = y^k + \frac{\roundbr{y^{k+1} - y^k}\cornerbr{s^k \xi^2 + \delta^k \xi(1 - \xi)}}{s^k + \cornerbr{\delta^{k+1} + \delta^k - 2s^k} \xi(1-\xi)}
\end{equation}
defined by the $k^{th}$ bin can be evaluated. For the inverse transformation, derivatives to compute the Jacobian determinant and further details, we refer the reader to~\cite{durkan2019neural}.

In practice, the knot coordinates and derivatives come from the conditioner network. Its output $\bm{\theta}_i \medeq \cornerbr{\bm{\theta}_i^w, \bm{\theta}_i^h, \bm{\theta}_i^d}$ is simply partitioned into vectors of length $K$, $K$ and $K-1$ for the knot widths and heights, as well as the knot derivatives. To compute the actual knot coordinates, $\bm{\theta}_i^w$ and $\bm{\theta}_i^h$ are softmax normalized, multiplied by $2B$ and their cumulative sums starting from $-B$ are computed.

\begin{figure}[t]
    \centering
    \includegraphics[width=0.42\textwidth]{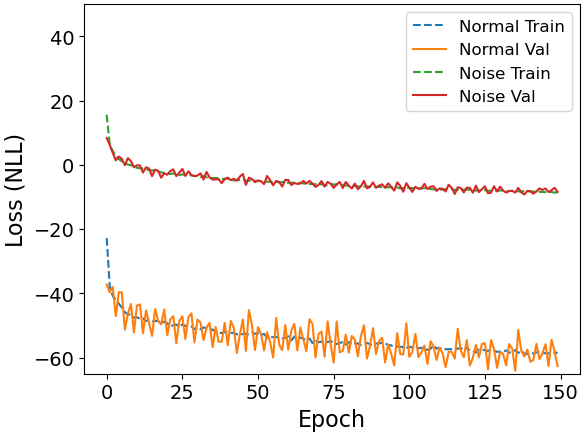}
    \caption{Comparison between normally training our model and with our proposed noise injection. The training becomes more stable and likelihoods stay in a reasonable range.}
    \label{fig:noise-injection}
\end{figure}

Finally, the sampled output of our model (after the last spline transformation) represents the predicted trajectory as relative displacements. As for using relative coordinates in the motion encoding~(see~\sect{sec:motionencoder}), this has proven to be beneficial for motion prediction~\cite{scholler2020constant}, and it also limits the numeric range of the output. This is important to stay within the support $\cornerbr{-B, B}$ of the spline transformations. We denote this estimated relative displacements as $\hat{\pred}^\prime$. To convert it back to absolute coordinates, we compute the cumulative sum over all positions $\pos \in \hat{\pred}^\prime$, starting from the last observed position $\pos^t$ of $\obs$. Because both the relative observed trajectory $\obs^\prime$ and the relative future trajectory $\pred^\prime$ can be unambiguously mapped back to $\obs$ and $\pred$, it holds that $p(\pred^\prime \g \obs^\prime) \medeq p(\pred \g \obs)$. Hence we learn the target distribution.

Furthermore, like in~\cite{chang2019argoverse}, before making a prediction we rotate the trajectory of the target agent around $\pos^t$, such that the last relative displacement $\pos^t - \pos^{t-1}$ is aligned with the vector $(1, 0)$. After sampling from our model, we rotate the predicted trajectories back. This transformation simplifies the distribution our model must learn and makes it rotation invariant. Because rotations are volume preserving, we do not have to consider this in our flow's likelihood computation.

\subsection{Preventing Manifolds}
\label{sec:manifolds}
Whenever data is distributed such that it --~or a subset of it~-- is residing on a lower-dimensional manifold, this leads to infinite likelihood spikes in the estimated density. Consider the two-dimensional example with joint density $p(x,y)$, where $x$ is normally distributed and $y=x$. The distribution resides on a line and for ${\int\int p(y\g x) p(x) \mathrm{d}y\mathrm{d}x=1}$ to hold, the likelihoods where $y$ is defined must be infinite.

In practice, this problem also arises when certain dimensions in the dataset samples frequently take on equal values, or when one dimension frequently takes the same value. Because we predict relative displacements $\pred^\prime$ instead of absolute coordinates, this can happen if pedestrians stand still (values become zero), or if they move with constant velocity for multiple timesteps (values are equal). During training this can cause numerical instabilities, loss volatility and the overestimation of certain samples' likelihoods.

To mitigate this problem and inspired by~\cite{kim2020softflow}, we define three hyper-parameters $\alpha$, $\beta$ and $\gamma$. While training, when transforming $\pred^\prime$ to $\bfu$ through the inverse of our flow, we augment $\pred^\prime$ before our first flow module as follows:
\begin{equation}\begin{split}
    \pred^{\prime\prime} &= \alpha \pred^\prime \\
    \predelem^{\prime\prime}_i &= \predelem^\prime_i + \epsilon_{\beta i} \quad \text{for all} \quad \predelem^\prime_i = 0 \\
    \predelem^{\prime\prime}_i &= \predelem^\prime_i + \epsilon_{\gamma i} \quad \text{for all} \quad \predelem^\prime_i \neq 0.
\end{split}\end{equation}
We sample noise vectors $\bm{\epsilon}_\beta$ and $\bm{\epsilon}_\gamma$ from zero-centered normal distributions with standard deviation $\beta$ and $\gamma$, respectively. However, we only apply noise during the training phase and not at inference time. In the forward pass, we always compute $\pred^\prime \medeq \alpha^{-1} \pred^{\prime\prime}$ after our last flow module to normalize predicted trajectories. By adding the noise during training, we essentially lift data off potential manifolds. Generally speaking, we apply less noise to zero-valued dimensions and more to non-zero displacement vectors. Scaling $\pred^\prime$ with $\alpha$ allows us to inject more noise, while controlling the impact of the noise on the trajectory.

The lower training curves in~\fig{fig:noise-injection} show how the loss of our model behaves when trained normally, without our noise injection. The loss is very volatile, especially for the validation dataset, and the likelihoods produced by our model are very large. Because we use the negative log likelihood loss (see~\sect{sec:loss-function}), these large likelihoods lead to an artificially low overall loss. However, empirically these inflated likelihoods do not correlate with better prediction performance and are meaningless. The upper curves in~\fig{fig:noise-injection} show how the training behaves with our noise injection. The magnitudes of the likelihoods are significantly reduced, because samples that originally lied on manifolds get smaller likelihood values assigned. Hence, they stop to dominate the training and this reduces the volatility of our validation loss. With our method, we experienced more reliable convergence during our experiments. Furthermore, it helps to avoid numerical problems during training and makes the model's likelihoods easier to use in downstream tasks (e.g. those that require normalization with softmax).

\subsection{Objective Function}
\label{sec:loss-function}
Because our model makes it easy to compute likelihoods for training examples (see~\eq{eq:flowinverse}), we simply train it with maximum likelihood estimation. In particular, we minimize the negative log likelihood
\begin{equation}
    \operatorname{NLL} = - \frac{1}{N} \sum_{i=1}^N {\log}(p(\pred_i \g \obs_i)).
\end{equation}

\subsection{Trajectory Augmentation}
To increase the diversity of our data, we augment trajectories by randomly scaling them. In particular, for each trajectory we sample a scalar in range $[s_{\min},s_{\max}]$ from a truncated normal distribution. Before multiplying the trajectory element-wise with the scalar, we first center the trajectory by subtracting its mean position to avoid translating it with the scaling, and then move it back. Scaling a trajectory does not influence its direction and motion pattern, but simulates varying movement speeds. It is crucial to stay within realistic limits by applying this transformation and the correct choice for the sampling interval depends on the used data.

%% file: sections/experiments.tex
\renewcommand{\arraystretch}{1.3}
\begin{table*}[t]
\begin{center}
\begin{tabular}{c|c|c|c|c|c|c}
\textbf{Model} & \textbf{ETH-Uni} & \textbf{Hotel} & \textbf{UCY-Uni}  &  \textbf{Zara1} & \textbf{Zara2} & \textbf{AVG} \\
\hhline{=======}
SGSG~\cite{xue2020scene} & 0.54 / 1.07 & 0.24 / 0.45 & 0.57 / 1.19 & 0.35 / 0.79 & 0.28 / 0.59 & 0.40 / 0.82 \\ \cline{1-7}
S-STGCNN~\cite{mohamed2020social} & 0.64 / 1.11 & 0.49 / 0.85 & 0.44 / 0.79 & 0.34 / 0.53 & 0.30 / 0.48 & 0.44 / 0.75 \\ \cline{1-7}
S-GAN~\cite{gupta2018social} & 0.59 / 1.04 & 0.38 / 0.80 & 0.27 / 0.49 & 0.18 / 0.33 & 0.19 / 0.35 & 0.32 / 0.60 \\ \cline{1-7}
CVM-S~\cite{scholler2020constant} & 0.44 / 0.81 & 0.20 / 0.35 & 0.34 / 0.71 & 0.25 / 0.49 & 0.22 / 0.45 & 0.29 / 0.56 \\ \cline{1-7}
Trajectron++~\cite{salzmann2020trajectron++} & 0.39 / 0.83 & \textbf{0.12} / \textbf{0.21} & \textbf{0.20} / \textbf{0.44} & \textbf{0.15} / \textbf{0.33} & \textbf{0.11} / \textbf{0.25} & \textbf{0.19} / 0.41 \\ \cline{1-7}
$\text{TF}_q$~\cite{giuliari2020trajtransformer} & 0.61 / 1.12 & 0.18 / 0.30 & 0.35 / 0.65 & 0.22 / 0.38 & 0.17 / 0.32 & 0.31 / 0.55 \\ \cline{1-7}
\textbf{FloMo (ours)} & \textbf{0.32} / \textbf{0.52} & 0.15 / 0.22 & 0.25 / 0.46 & 0.20 / 0.36 & 0.17 / 0.31 & 0.22 / \textbf{0.37} \\
\hhline{=======}
\end{tabular}
\end{center}
\caption{Displacement errors for scenes in the ETH/UCY datasets and on average. We compare our model to six state-of-the-art models. Each model predicted 20 trajectory samples and the errors are shown as minADE / minFDE.}
\label{tab:errors-ethucy}
\end{table*}

We evaluate our model with the publicly available ETH\cite{pellegrini2009eth}, UCY~\cite{lerner2007ucy} and Stanford Drone~\cite{robicquet2016learning} motion datasets. All datasets are based on real-world video recordings and contain complex motion patterns. The ETH/UCY datasets are evaluated jointly and focus on pedestrians that were recorded in city centers and at university campuses. They cover a total of five distinct scenes with four unique environments and 1950 individual pedestrians. The larger Stanford Drone dataset contains 10300 individual traffic participants, it covers roads and besides pedestrians it includes also other agent types like cyclists and vehicles. All datasets are heavily used in the motion prediction domain~\cite{alahi2016social, scholler2020constant, salzmann2020trajectron++, bhattacharyya2020haar, sadeghian2018sophie, gupta2018social}.

\renewcommand{\arraystretch}{1.3}
\begin{table}[htpb]
\begin{center}
\begin{tabular}{c|c|c|c|c}
\textbf{Model} & \textbf{@\SI{1}{\second}} & \textbf{@\SI{2}{\second}} & \textbf{@\SI{3}{\second}} & \textbf{@\SI{4}{\second}} \\
\hhline{=====}
STCNN~\cite{pajouheshgar2018back} & 1.20 & 2.10 & 3.30 & 4.60  \\ \cline{1-5}
FlowWaveNet~\cite{kim2018flowavenet}\cite{bhattacharyya2020haar} & 0.70 & 1.50 & 2.40 & 3.50  \\ \cline{1-5}
HBA-Flow~\cite{bhattacharyya2020haar} & 0.70 & 1.40 & 2.30 & 3.20  \\ \cline{1-5}
\textbf{FloMo (ours)}  & \textbf{0.27} & \textbf{0.56} & \textbf{0.90} & \textbf{1.27} \\ 
\hhline{=====}
\end{tabular}
\end{center}
\caption{Errors for the Stanford Drone dataset, evaluated with a five-fold cross-validation, the Oracle Top 10\% metric and 50 predicted trajectories. All models are tractable and allow exact likelihood computation.}
\label{tab:errors-stanford-1}
\end{table}

\renewcommand{\arraystretch}{1.3}
\begin{table}[htpb]
\begin{center}
\begin{tabular}{c|c|c}
\textbf{Model} & \textbf{minADE} & \textbf{minFDE}\\
\hhline{===}
SocialGAN~\cite{gupta2018social} & 27.23 & 41.44  \\ \cline{1-3}
SoPhie~\cite{sadeghian2018sophie} & 16.27 & 29.38  \\ \cline{1-3}
CF-VAE~\cite{bhattacharyya2019conditional} & 12.60 & 22.30  \\ \cline{1-3}
HBA-Flow~\cite{bhattacharyya2020haar} & 10.80 & 19.80 \\ \cline{1-3}
PECNet~\cite{mangalam2020not} & 9.96 & 15.88 \\ \cline{1-3}
\textbf{FloMo (ours)}  & \textbf{2.60} & \textbf{4.43} \\ 
\hhline{===}
\end{tabular}
\end{center}
\caption{Evaluation results for Stanford Drone with a single dataset split, 20 predicted trajectories and the minADE / minFDE metrics. Here we also include intractable models.}
\label{tab:errors-stanford-2}
\end{table}
We follow for all datasets the most common evaluation regimes. For the ETH/UCY datasets we always train on four scenes and evaluate on the remaining one. We slice each trajectory with a step-size of one into sequences of length 20, of which 8 timesteps are observed and 12 must be predicted. This corresponds to an observation window of \SI{3.2}{\second} and a prediction of \SI{4.8}{\second}. For the Stanford Drone dataset we randomly split into training and testset but ensure that both sets do not contain parts of the same video sequences. We observe for 20 timesteps and predict the next 40 timesteps, which corresponds to \SI{2}{\second} and \SI{4}{\second}, respectively. For comparability, we follow~\cite{lee2017desire, bhattacharyya2020haar} and scale the dataset trajectories by a factor of $\nicefrac{1}{5}$.

For training our model, we only take into account trajectories of full length, because padding would cause issues as described in~\sect{sec:manifolds}. However, in our evaluation we use all trajectories that have a length of at least 10 timesteps for ETH/UCY, and at least 22 timesteps for Stanford Drone, i.e. at least two timesteps to predict. Note that we also compare tractable models only based on displacement errors and not on log likelihoods. While our model's likelihoods are meaningful, as we show in~\sect{sec:likelihoods}, the overall log likelihood for trajectory datasets is largely dominated by manifold artifacts and hence not ideal for comparison. \\

\vspace{-0.6em}

\textbf{Training.} We trained our model with the Adam~Optimizer~\cite{kingma2014adam}, learning rate 0.001, and batch size 128 for 150 epochs. We randomly split a 10\% validation set for ETH/UCY and a 5\% validation set for Stanford Drone from each training set to detect overfitting. Furthermore, we define the support for each spline flow as $B\tighteq15$ and use 8 knot points. For ETH/UCY we set $\alpha\tighteq10$, $\beta\tighteq0.2$, $\gamma\tighteq0.02$ and for Stanford Drone $\alpha\tighteq3$, $\beta\tighteq0.002$, $\gamma\tighteq0.002$. In our scaling transformation we set $\mu\tighteq1$ for all datasets, but for ETH/UCY $\sigma=0.5$, $s_{\min}\tighteq0.3$, $s_{\max}\tighteq1.7$ and for Stanford Drone $\sigma\tighteq0.2$, $s_{min}\tighteq0.8$, $s_{max}\tighteq1.2$. In total, we stack 10 flow layers in our model. All hyper-parameters described were determined empirically.

\vspace{0.5em}

\textbf{Metrics.} As proposed by~\cite{gupta2018social}, we allow each model to predict multiple samples. For the ETH/UCY datasets we report errors in meters, and for the Stanford Drone dataset in pixels. We evaluate with the following metrics:

\begin{itemize}
    \item \textit{Minimum Average Displacement Error (minADE)} --- Error of the sample with the smallest average L2 distance between all corresponding positions in the ground truth and the predicted trajectory.
    \item \textit{Minimum Final Displacement Error (minFDE)} --- Error of the sample with the smallest L2 distance between the last position in the ground truth and the last position in the predicted trajectory.
    \item \textit{Oracle Top 10\%} --- Average error of the top 10\% best predicted trajectories at different timesteps. It has been shown that this measure is robust to random guessing and simply increasing the number of drawn samples does not affect it~\cite{bhattacharyya2019conditional}.
\end{itemize}

\textbf{Baselines.} We compare our model with a variety of state-of-the-art prediction models. Except the \mbox{CVM-S}~\cite{scholler2020constant}, all other models are based on neural networks. \mbox{S-STGCNN}~\cite{mohamed2020social}, SGSG~\cite{xue2020scene} and Trajectron++~\cite{salzmann2020trajectron++} utilize neural networks in combination with graphs. $\text{TF}_q$~\cite{giuliari2020trajtransformer} is based on the transformer architecture. \mbox{S-GAN}~\cite{gupta2018social}, \mbox{SoPhie}~\cite{sadeghian2018sophie} are GANs. STCNN~\cite{pajouheshgar2018back}, FloWaveNet and HBA-Flow are exact inference models and the latter two based on normalizing flows. Besides the Trajectron++, also CF-VAE\cite{bhattacharyya2019conditional} and PECNet~\cite{mangalam2020not} use a conditional VAE as their core network.

\begin{figure}[t]
    \centering
    \includegraphics[width=0.39\textwidth]{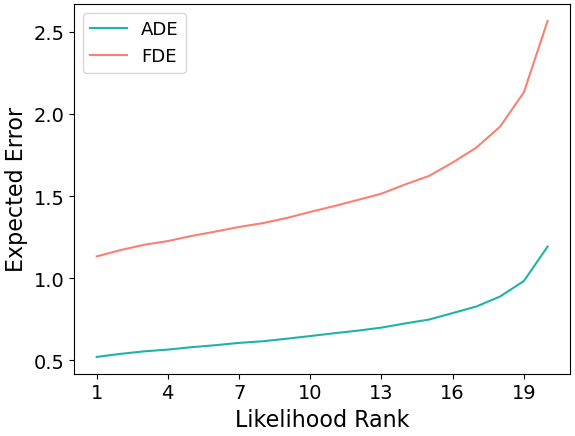}
    \caption{Relationship between our model's prediction errors and associated likelihood ranks for the ETH/UCY datasets.}
    \label{fig:likelihooderror}
\end{figure}

\begin{figure}[t]
    \centering
    \begin{subfigure}[b]{0.44\textwidth}
        \includegraphics[trim=0 60 0 0,clip,width=1.0\textwidth]{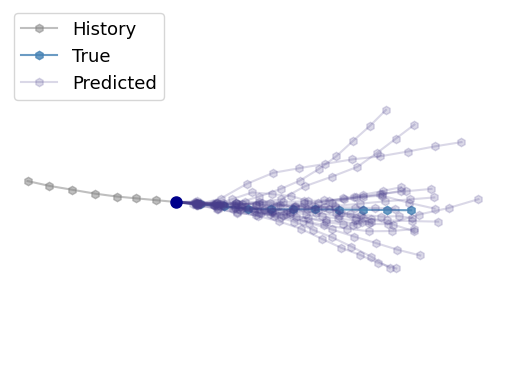}
        \caption{Regular predictions}
        \label{fig:predictions-a}
    \end{subfigure}
    \par\medskip
    \begin{subfigure}[b]{0.44\textwidth}
        \includegraphics[trim=0 80 0 60,clip,width=1.0\textwidth]{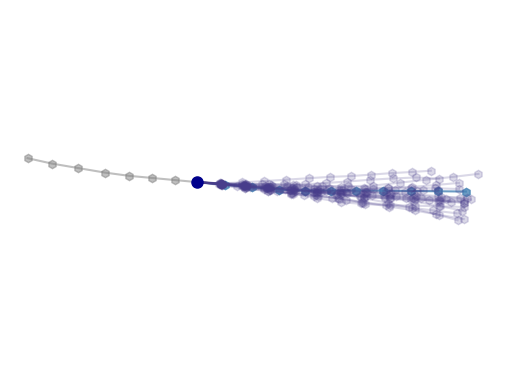}
        \caption{Top-k predictions}
        \label{fig:predictions-b}
    \end{subfigure}
    \caption{Comparison of our model's regular predictions with top-k predictions for the same sample.}
    \label{fig:predictions}
\end{figure}

\subsection{Displacement Errors}
\label{sec:errors}
For the ETH/UCY datasets, we compare our model with state of the art in~\tab{tab:errors-ethucy}. Following the standard protocol, each model was allowed to predict 20 trajectory samples in this evaluation. Except for the \mbox{Trajectron++}, our model significantly outperforms all other models on average errors, both in terms of minADE and minFDE. Compared to the \mbox{Trajectron++}, our model performs better on the ETH-Uni scene, while on the other Scenes the \mbox{Trajectron++} achieves lower errors, especially for minADE. However, for the minFDE both models perform close on all scenes except ETH-Uni and Zara2. In total, the \mbox{Trajectron++} achieves lower errors averaged over the whole trajectories with a minADE of 0.19, but FloMo performs better on the endpoint prediction where it achieves a minFDE of 0.37. Hence, the prediction performance of both models can be considered as approximately equivalent. However, unlike the \mbox{Trajectron++} our model is tractable and allows direct likelihood computation. The close performance of both models could indicate that the noise floor for ETH/UCY predictions is approached.

On the Stanford Drone dataset we evaluated with two different protocols. For the results in~\tab{tab:errors-stanford-1} we performed a five-fold cross-validation and let each model predict 50 samples. Then we evaluated with the Oracle Top 10\% metric that we described earlier. All models in this evaluation allow tractable likelihood computation, and the concurrently proposed models \mbox{HBA-Flow} and \mbox{FlowWaveNet} (applied to motion prediction by~\cite{bhattacharyya2020haar}) are also based on normalizing flows. The displacement errors are evaluated at four different timesteps. Our model significantly outperforms all other models at each timestep with an improvement of 60\% at \SI{4}{\second} over the second best model HBA-Flow. This results show that our model captures the true underlying distribution better than the other tractable models.

In~\tab{tab:errors-stanford-2} we performed a second evaluation on the Stanford Drone dataset with a single dataset split, 20 trajectory predictions, and the minADE and minFDE metrics. In this case we also compare to intractable models. The results on this experiment confirm those of the previous experiment. Our model significantly outperforms all compared models, with a margin of 74\% in minADE and 72\% in minFDE compared to the second best model PECNet.

\subsection{Likelihoods}
\label{sec:likelihoods}
To verify that the likelihoods our model provides are relevant, we rank each of the 20 trajectory samples generated by our model for the ETH/UCY datasets in descending order by likelihood. Then we compute the expected ADE and FDE for each likelihood ranking position across all testsets. As for the evaluation in the previous section, for each testset evaluation we use the FloMo trained on the remaining scenes. \fig{fig:likelihooderror}~shows graphs of how the expected errors change with likelihood ranking. As expected, a higher likelihood (lower rank) corresponds to lower errors for both ADE and FDE. This proves that the likelihoods computed by our model are meaningful and can be used for decision making.

To qualitatively demonstrate how likelihoods relate to the predicted trajectories, in~\fig{fig:predictions-a} we show 20 regularly predicted trajectories and in~\fig{fig:predictions-b} a top-k prediction for the same example. For the top-k prediction we sample 100 trajectory candidates and only keep the 20 most likely ones. The regular predictions are much more spread out. Our model predicts sudden turns, acceleration, or deceleration. The top-k predictions are more concentrated around the true and most likely outcome of the pedestrian's movement. Furthermore, the predicted velocities are more regular. This results demonstrate that an autonomous agent can utilize the likelihoods our model provides to decide which predictions it should prioritize in its planning. 

\subsection{Ablation}
\renewcommand{\arraystretch}{1.3}
\begin{table}[t]
\begin{center}
\begin{tabular}{c|c|c}
\textbf{Method} & \textbf{ETH/UCY} & \textbf{Stanford Drone}\\
\hhline{===}
No Scaling & 0.27 / 0.46  & 2.92 / 5.02 \\ \cline{1-3}
Scaling & \textbf{0.22} / \textbf{0.37} & \textbf{2.60} / \textbf{4.43}  \\ \cline{1-3}
\hhline{===}
\end{tabular}
\end{center}
\caption{Ablation results for our scaling transformation.}
\label{tab:ablation}
\end{table}

To understand the impact of our scaling transformation on our model's performance, we conducted an ablation study. The results of this study for the ETH/UCY and the Stanford Drone datasets are shown in~\tab{tab:ablation}. Applying our transformation improved our model's performance on all datasets. By simulating varying movement speeds and thus diversifying the training data, our model learned to generalize better. We also analyzed our noise injection and found that it does not have a significant impact on average prediction performance. Most likely because the inflated density points are sparsely distributed. However, the injection's stabilizing effect on the training of our model, along with its numerical and practical advantages, make it a useful tool for training flows for motion prediction.

%% file: sections/conclusion.tex
In this work we proposed a motion prediction model based on spline flows that is able to learn a distribution over the future motion of agents. It makes it possible to directly compute likelihoods that are necessary for autonomous agents to prioritize predictions. Because training on trajectory data directly causes loss volatility and numerical instabilities, we proposed a method of injecting noise, such that training is stabilized, but the motion information in the trajectories is preserved. Furthermore, we suggested an augmentation transformation that improves our model's generalization.

To evaluate our model we conducted extensive experiments, in which we showed that our model achieves state-of-the-art performance in terms of displacement errors. We also showed at a quantitative and qualitative level that the likelihoods our model provides are meaningful and can be used for decision making in autonomous agents. With an ablation study, we ensured that our data augmentation transformation contributes positively to our model's performance.